\documentclass{article}

\usepackage{arxiv}

\usepackage[utf8]{inputenc} 
\usepackage[T1]{fontenc}    
\usepackage{hyperref}       
\usepackage{url}            
\usepackage{booktabs}       
\usepackage{amsfonts}       
\usepackage{nicefrac}       
\usepackage{microtype}      
\usepackage{lipsum}
\usepackage{authblk}

\usepackage[sort&compress,square,comma,numbers]{natbib}

\usepackage{enumitem}

\usepackage{float} 
\usepackage{times}
\usepackage{epsfig}
\usepackage{graphicx}
\usepackage{amsmath, amsfonts}
\usepackage{bm}
\usepackage{algorithm}
\usepackage{algorithmicx}
\usepackage{comment}
\usepackage{fullpage}
\usepackage{tikz}
\usepackage{pifont}
\usepackage{amsthm}
\usepackage{subcaption}
\newtheorem{definition}{Definition}

\newtheorem*{remark}{Remark}
\newtheorem*{example}{Example}

\DeclareMathOperator*{\argmin}{arg\,min}

\usepackage{mathtools}
\usepackage{multirow}
\usepackage[noend]{algpseudocode}
\usepackage{booktabs}
\usepackage[export]{adjustbox}
\usepackage{url}

\usepackage[textsize=scriptsize,backgroundcolor=green]{todonotes}

\providecommand{\keywords}[1]{\textbf{\textit{Index terms---}} #1}

\usepackage{xspace}

\newcommand*{\st}{\textit{s.t.}\@\xspace}

\makeatletter
\newcommand*{\etc}{%
	\@ifnextchar{.}%
	{\textit{etc}}%
	{\textit{etc.}\@\xspace}%
}
\makeatother
\algtext*{EndWhile}
\algtext*{EndFor}
\algtext*{EndIf}
\algdef{SE}[DOWHILE]{Do}{doWhile}{\algorithmicdo}[1]{\algorithmicwhile\ #1}%

\makeatletter
\def\BState{\State\hskip-\ALG@thistlm}
\makeatother

\title{Tail-Net: Extracting Lowest Singular Triplets for Big Data Applications}

\author[1]{\textbf{Gurpreet Singh} \textsuperscript{\dag}}
\author[2]{\textbf{Soumyajit Gupta} \textsuperscript{\dag}}
\affil[2]{Department of Computer Science}
\affil[1]{The University of Texas at Austin}
\affil[ ]{\texttt{\{gurpreet, smjtgupta\}@utexas.edu}}

\begin{document}

\maketitle

{\let\thefootnote\relax\footnote{{\dag contributed equally to this work.}}}

\begin{abstract}

 SVD serves as an exploratory tool in identifying the dominant features in the form of top rank-r singular factors corresponding to the largest singular values. For Big Data applications it is well known that Singular Value Decomposition (SVD) is restrictive due to main memory requirements. However, a number of applications such as community detection, clustering, or bottleneck identification in large scale graph data-sets rely upon identifying the lowest singular values and the singular corresponding vectors. For example, the lowest singular values of a graph Laplacian reveal the number of isolated clusters (zero singular values) or bottlenecks (lowest non-zero singular values) for undirected, acyclic graphs. A naive approach here would be to perform a full SVD however, this quickly becomes infeasible for practical big data applications due to the enormous memory requirements. Furthermore, for such applications only a few lowest singular factors are desired making a full decomposition computationally exorbitant. In this work, we trivially extend the previously proposed Range-Net to \textbf{Tail-Net} for a memory and compute efficient extraction of lowest singular factors of a given big dataset and a specified rank-r. We present a number of numerical experiments on both synthetic and practical data-sets for verification and bench-marking using conventional SVD as the baseline. 
 

\end{abstract}

\section{Introduction}

For low rank approximation, Singular Value Decomposition (SVD) is used to identify the top rank-r factors corresponding to the largest singular values of a given data matrix $X$. For big data applications, Range-Net \cite{singh2021range} was proposed as a memory efficient and accurate alternative to conventional and randomized SVD. However, for applications such as community detection, clustering, or identification of connectivity bottlenecks from large scale graph data-sets (adjacency lists or matrices) require efficiently extracting lowest singular factors. We present an extension of Range-Net as \textbf{Tail-Net} that again has an exact memory requirement and accurately extracts the lowest singular factors without performing a full SVD. In the following we first describe the modified problem statement (compare to Range-Net \cite{singh2021range}) and translate this to the originally proposed Range-Net. Following Eckart-Young-Mirsky theorem \cite{eckart1936approximation}, we first define a \textbf{b-tail energy} for extracting the lowest singular factors analogous to the original definition of the tail energy described in this seminal work equipped with a Frobenius norm. As before, we exploit the properties of Frobenius norm and this b-tail energy to define an appropriate network loss function. Since this formulation is a minor deviation from the original Range-Net, a number of components are either reiterated or adapted from our previous work.

\begin{figure}[h]
    \centering
    \includegraphics[width=\linewidth]{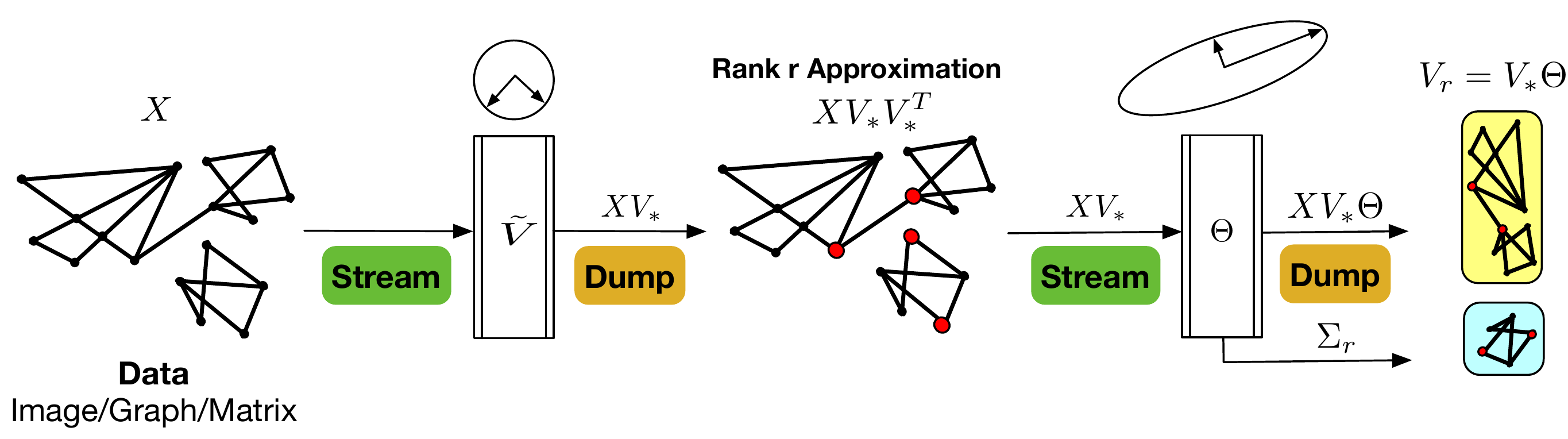}
    \caption{An overview of the low-memory, two-stage Tail-Net SVD for Big Data Applications. Stage 1 identifies the span of the desired rank-r approximation. Stage 2 rotates this span to align with the singular vectors while extracting the singular values of the data. The input data can be streamed from either a server or secondary memory. The number of zero singular values of a graph Laplacian indicate the number of isolated clusters with the lowest singular triplets indicative of bottlenecks (red dots).}
    \label{fig:overview}
    \vspace{-2mm}
\end{figure}

\textbf{Fig. \ref{fig:overview}} shows an overview of the proposed Tail-Net's two-stage rank-r approximation corresponding for extracting the lowest energy factors. As described below, this two stage approach remains the same as in the case of the originally proposed Range-Net with minor changes in the stage-1 loss function following the now modified problem statement.

\subsection{Problem Statement}\label{sec:prob}

Let us denote the data matrix as $X \in \mathbb{R}^{m \times n}$ of rank $f \leq \min(m,n)=g$ and its approximation as $X_{r} \in \mathbb{R}^{m \times n}$, where $g = min(m,n)$ for convenience in notation. The singular value decomposition of $X = U\Sigma V^{T}$, where $U \in \mathbb{R}^{m \times n} = [u_1, \cdots, u_f]$ and $V \in \mathbb{R}^{n \times f} = [v_1, \cdots, v_f]$ are its left and right singular vectors respectively, and $\Sigma \in \mathbb{R}^{f \times f} = diag(\sigma_1,\cdots,\sigma_f)$ are the corresponding non-zero, singular values. The lowest rank $r$ truncation of $X$ is then $X_r = U_r \Sigma_r V_r^T$, where $\Sigma_{r}$ is a diagonal matrix of the smallest $r$ singular values of $X$, and $U_{r} = U_{[g-r:g]}$ and $V_{r}=V_{[g-r:g]}$ are the corresponding left and right singular vectors. In other words, $X = U\Sigma V^T=U_r\Sigma_r V^T_r+U_{g \textbackslash r}\Sigma_{g \textbackslash r}V^T_{g \textbackslash r} = X_{r} + X_{g \textbackslash r}$. Here, $U_{g \textbackslash r},V_{g \textbackslash r}$ are the leading $g-r$ left and right singular vectors, respectively.

\begin{definition}
Following the seminal work of Eckart-Young-Mirsky \cite{eckart1936approximation,mirsky1960symmetric}, for a top rank-$r$ approximation of a given data matrix $X$ with the tail energy defined as $\|X-X\tilde{V}\tilde{V}^{T}\|_{F}$, we define \textbf{b-tail energy} as $\|X\tilde{V}\tilde{V}^{T}\|_{F}$.
\end{definition}

From here forth, this modified tail-energy will be referred to as the b-tail energy with the intent of extracting the bottom rank-$r$ singular factors. The problem statement is then: Given $X \in \mathbb{R}^{m \times n}$ find $\tilde{V}\in \mathbb{R}^{n \times r}$ such that,
\begin{equation}
    \argmin \limits_{\substack{\tilde{V}\in \mathbb{R}^{n \times r} \\ \mathrm{rank}(X\tilde{V}\tilde{V}^{T}) \leq r}} \|X \tilde{V} \tilde{V}^{T} \|_{F} \quad \text{\st} \quad \tilde{V}^{T}\tilde{V}=I_{r}
    \label{eq:eym1}
\end{equation}
where $\|\cdot\|_{F}$ is the Frobenius norm. As in the case of Range-Net, the minimizer $\tilde{V}_{*}$ naturally gives the lower bound on this b-tail energy that satisfying the aforementioned constraint. Note that in the absence of this constraint ($\tilde{V}^{T}\tilde{V}=I_{r}$), we arrive at trivial solutions corresponding to $\tilde{V}=[v_1,\cdots,v_r], v_i \in null(X)$ $\forall$ $i=1,\cdots,r$. In the following, we trivially modify one term in the Stage-1 loss function of Range-net without changing Stage-2 giving rise to \textbf{Tail-Net} that now extracts the lowest rank-$r$ factors of a given data matrix $X$ (sparse or dense).
\begin{remark}
Note that under the above definition of truncated SVD, the left and right singular vectors span a null space of size $(g-f)$. Therefore, if $r\leq (g-f)$ then all $r$ singular values will be zero, whereas if $r > (g-f)$ then we have $r-(g-f)$ non-zero singular values with the remaining $(g-f)$ zero singular values.
\end{remark}
\begin{example}
Let us consider a square matrix $X \in \mathbb{R}^{10 \times 10} = diag(7,6,5,4,3,2,1,0,0,0)$ where, $m=n=10$ and $f=7$. If $r=2$, then $\sigma_{i=1,2}=0$ or if $r=5$, then $\sigma_{i=1,2,3}=0$ and $\sigma_4=1,\sigma_5=2$.
\end{example}

\subsection{Main Contributions}

\textbf{Minimal Assumptions:} The proposed approach does not make any assumptions on the rank or decay rate of singular values given a data matrix. For practical applications neither of these are known \textit{a priori} and therefore must be avoided at all times. The proposed approach only assumes that the matrix is positive semi-definite so that the singular values are either positive or zero.

\textbf{Data and Representation Driven Neural SVD:} The representation driven network loss terms ensures that the data matrix $X$ is decomposed into the desired SVD factors such that $X = U\Sigma V^{T}$. In the absence of the representation enforcing loss term, the minimizer of \textbf{Eq. \eqref{eq:eym1}} results in an arbitrary decomposition such that $X = ABC$ different from SVD factors.

\textbf{A Deterministic Approach with GPU Bit-precision Results:} The network is initialized with weights drawn from a random distribution while the iterative minimization is deterministic. Although not advised for big data matrices, a full gradient descent converges to the same minimizer as a stochastic gradient descent.

\textbf{Streaming Architecture with Exact Low Memory Cost:} Tail-Net requires an exact memory specification based upon the desired rank-$r$ and data dimensions $X \in \mathbb{R}^{m\times n}$ given by $r(n+r)$ and not $\mathcal{O}(r(n+r))$ independent of the sample dimension $m$. The streaming order of the samples is of no consequence and the user is free to choose the order in which the samples are processed in a batch-wise manner (indexed or randomized). 

\textbf{Layer-wise Fully Interpretable:} Tail-Net is a low-weight, fully interpretable, dense neural network where all the network weights and outputs have a precise definition and the choice of network activations is strict. The network weights are placeholders for the right (or left) orthonormal vectors upon convergence of the network minimization problems. The user can explicitly plug a ground truth solution to verify the network design and energy bounds. 


\section{Motivation}

For most dimension reduction and compression applications, we are primarily interested in extracting the top singular triplets. A trivial yet important question to ask is that: What is the need to extract the lowest singular triplets of any data matrix? Although not prominent, lowest singular triplets find a wide variety of applications including; detecting number of isolated communities in a graph \cite{von2007tutorial}; Slow Feature Analysis \cite{wiskott2002slow} to find the low frequency features in a time series; Minor Component Analysis \cite{luo1997minor} for  moving target indication and clutter cancellation; Extreme Component Analysis \cite{reiss1997statistical} for detecting extremes in financial time series and values at risk; Canonical Correlation Analysis \cite{hardoon2004canonical} for detecting private spaces between mixed datasets; minimizing residuals in Total Least Squares \cite{bjorck2000methods}; K-plane clustering \cite{bradley2000k} for Information Retrieval applications; compute the distance to uncontrollability \cite{elsner1991algorithm} for control systems; Pseudo-spectra of a matrix \cite{trefethen1999computation}; subspace signal analysis \cite{van1993subspace} \etc

\begin{figure}[h]
    \centering
    \begin{subfigure}{.45\linewidth}
      \centering
      \includegraphics[width=0.6\linewidth]{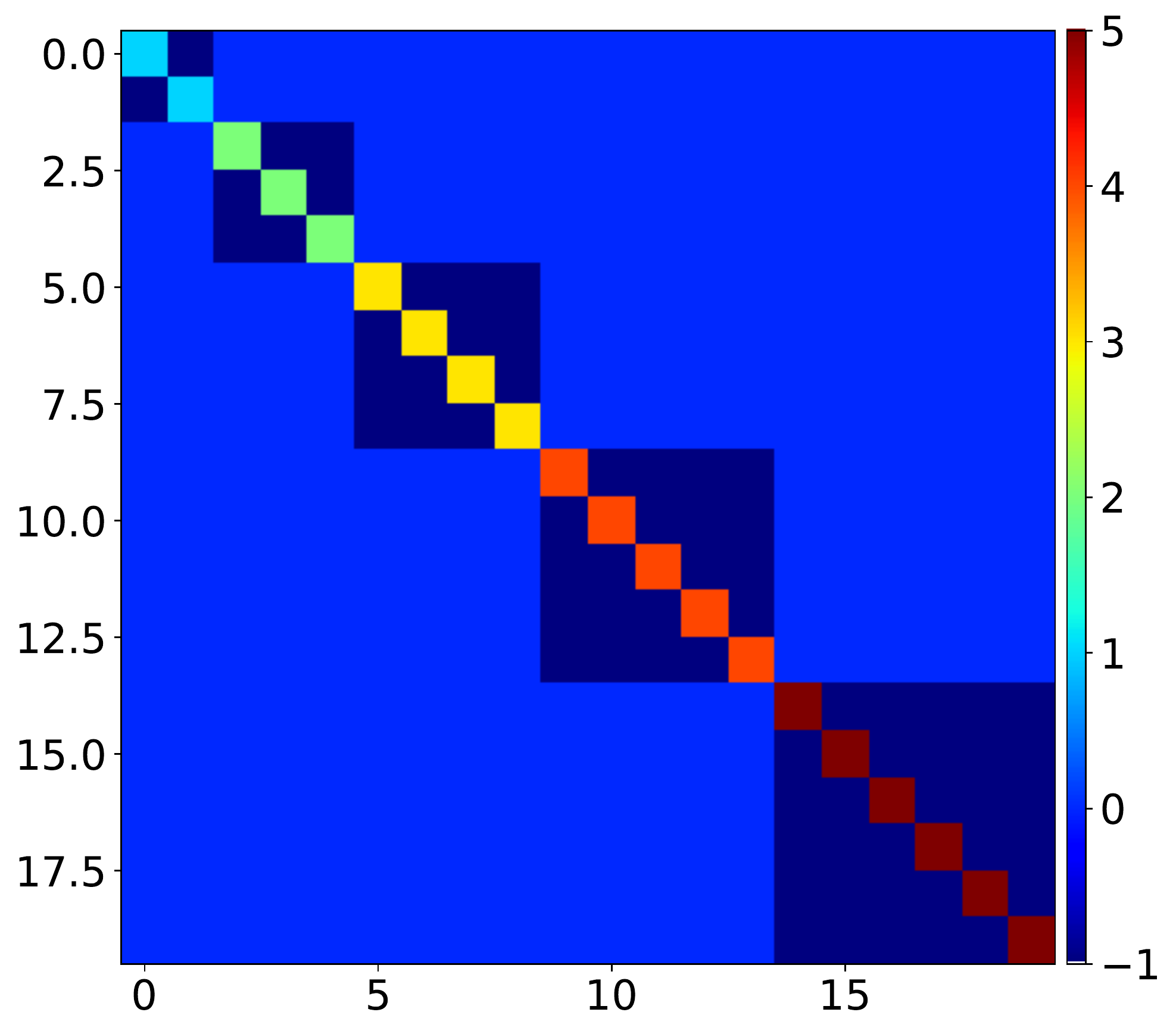}  
      \caption{Graph Laplacian}
    \end{subfigure}
    \begin{subfigure}{.45\linewidth}
      \centering
      \includegraphics[width=0.8\linewidth]{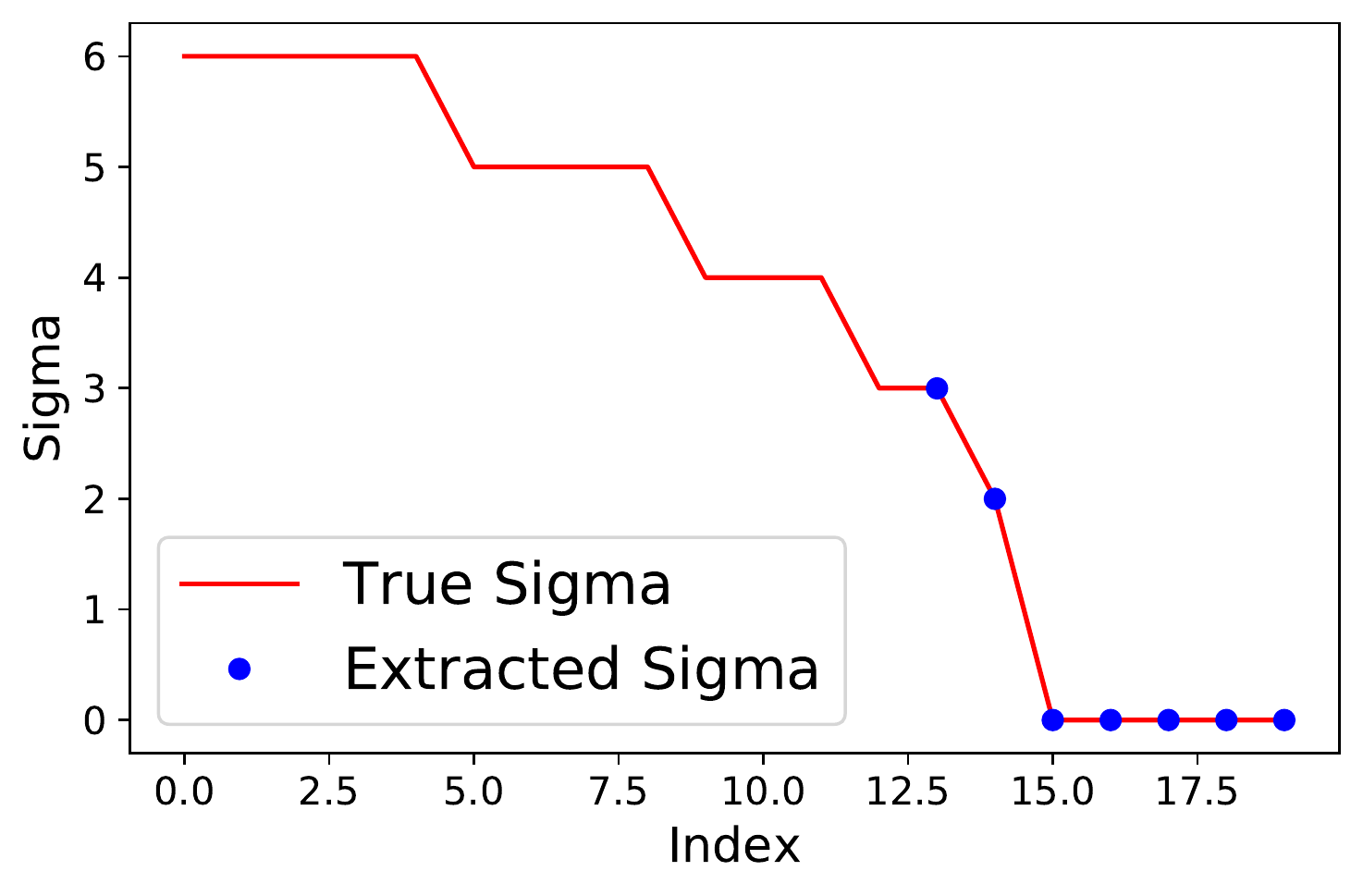}  
      \caption{Singular value spectrum}
    \end{subfigure}
    \caption{The synthetic adjacency matrix has $20$ nodes with $5$ clusters of varying sizes. (a) Shows the corresponding graph Laplacian matrix and (b) shows the true (in reverse order) and Tail-Net extracted singular values. The lowest $5$ singular values are zero corresponding to the five isolated communities. Tail-Net extracted values up to GPU bit precision matching the true values. Note the rank-revealing quality of Tail-Net.}
    \label{fig:comm}
\end{figure}

As a use-case scenario, we present Tail-Net for isolated community detection in graph processing applications. \textbf{Fig. \ref{fig:comm}} shows a synthetic case of a graph consisting of five communities of varying sizes. For the graph Laplacian \cite{von2007tutorial}, we expect the last $5$ singular values to be zero, followed by non-zero values. As show in Fig. \textbf{Fig. \ref{fig:comm}} (b), Tail-Net correctly identifies the seven values up to GPU precision with five zero and two non-zero singular values indicating the presence of five isolated clusters.

\section{Related Works}

A limited number of approaches have been proposed in the existing literature to extract the lowest rank-$r$ singular triplets accurately. Readers are referred to \citet{wu2014primme,dax2019computing}, and the citations therein, for a detailed review on the current state of the art. The most used ones that are deployed in software packages as \textit{svds} function in Matlab include \citet{larsen1998lanczos,baglama2005augmented}. These approaches rely upon constructing augmented matrices ($B$ and $C$ defined below) for extracting the lowest singular triplets. 

\begin{definition}
For a given data matrix $x \in \mathbb{R}^{m \times n}$, the augmented matrices $B$ and $C$ are defined as,
    \begin{align*}
        B \in \mathbb{R}^{(m+n) \times (m+n)}= \begin{bmatrix}
        0 & X^T\\ 
            X & 0
        \end{bmatrix} \qquad 
        C \in \mathbb{R}^{n \times n} = X^TX
    \end{align*}
\end{definition}

Although accurate, these approaches require a persistent presence of either the data or the augmented matrices in the main memory. As stated before, this becomes infeasible for big data applications since the peak memory load of these solution algorithms is often a multiplicative factor ($>1$) of the memory requirement for loading the dataset itself. This is further exacerbated when the data matrix is not sparse. Additionally, for rank-deficient matrices \cite{dax2019computing} requires shift and invert process so that the augmented matrix becomes invertible. Further, these approaches either extract singular values one by one in an increasing order or switch between multiple methods leaving little room to design algorithms to alleviate the memory load. Motivated by these limitations we propose \textbf{Tail-Net}, which similar to it's predecessor Range-Net, has an exact memory requirement dependent on the number of features in the dataset and the desired rank-r, independent of the number of samples. Additionally, Tail-Net does require that the data matrix be full rank and can handle low-rank matrices $\left(f\leq min(m,n)\right)$. In this respect, Tail-Net is rank revealing as will be shown later through numerical experiments. 

\begin{remark}
Note that, to date no randomized SVD algorithms, for reduced memory consumption, have been proposed to address this problem and therefore we do not discuss any of these methods here. The current problem now requires computing an accurate projector ($\tilde{V}_{*}\tilde{V}_{*}^{T}$) for extracting the lowest singular triplets of a data matrix $X$.
\end{remark}


\section{Tail-Net: A 2-stage Lowest Singular Triplet Extractor}

In the following, we present \textbf{Tail-Net} that explicitly relies upon solving the minimization problem in Eq. \eqref{eq:eym1} to achieve the lower bound on the b-tail energy for a desired rank-$r$ approximation of a data matrix $X$ under a streaming setting. The readers are referred to \cite{singh2021range} to familiarize themselves with the original Range-Net network architecture and loss functions.

\subsection{Network Architecture}\label{sec:arch}

\begin{figure}[h]
    \centering
    \includegraphics[width=0.6\linewidth]{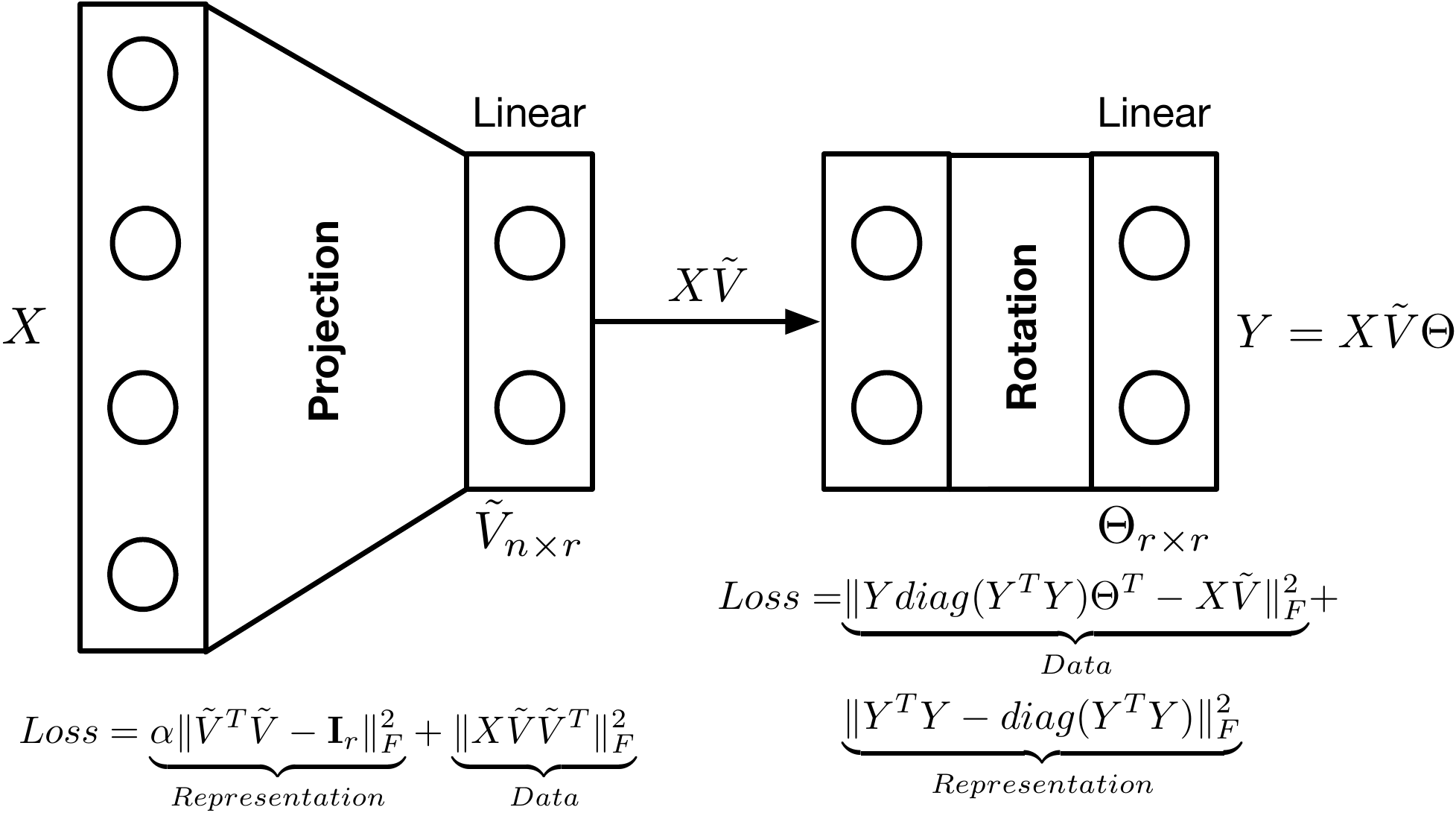}
    \caption{Tail-Net Architecture: Projection (Net1) and Rotation (Net2) for a 2-stage SVD. Compared to the original Range-Net only one loss term changes to extract the lowest singular factors.}
    \label{fig:arch}
\end{figure}

The proposed network architecture is divided into two stages: (1) Projection, and (2) Rotation, each containing only one dense layer of neurons and linear activation functions with no biases. \textbf{Fig. \ref{fig:arch}} shows an outline of the this two-stage network architecture where all the weights and outputs have a specific meaning enforced using representation and data driven loss terms. Contrary to randomized SVD algorithms the subspace projection (Stage 1) is not specified preemptively (consequently no assumptions) but is computed by solving an iterative minimization problem following EYM theorem corresponding to Eq. \eqref{eq:eym1}. The rotation stage (Stage 2) then reuses the EYM tail-energy to extract the singular vectors and values.

\begin{remark}
All network activations are chosen to be linear with no-biases following the arguments presented in \cite{singh2021range}.
\end{remark}

\textbf{Stage 1: Rank-$r$ Sub-space Identification:} The projection stage constructs an orthonormal basis that spans the $r$-dimensional sub-space of a data matrix $X \in \mathbb{R}^{m \times n}$ of an unknown rank $f \leq min(m,n) = g$. This orthonormal basis ($\tilde{V}$) is extracted as the stage-1 network weights once the network minimization problem converges to a fixed-point. The representation loss $\|\tilde{V}^T\tilde{V}-I_r\|_{F}$ in stage-1 enforces the orthonormality requirement on the projection space while the data-driven loss $\|X\tilde{V}\tilde{V}^T\|_{F}$ minimizes the b-tail energy.  Although the minimization problem is non-convex the b-tail energy is guaranteed to converge to the minimum, as in the case of Range-Net \cite{singh2021range}. The stage-1 minimization problem following the problem statement given Eq. \eqref{eq:eym1} then reads,
\begin{align}
    \underset{\tilde{V}}{\min} \quad \|X\tilde{V}\tilde{V}^{T}\|_{F} + \alpha \|\tilde{V}^{T}\tilde{V} - I_{r}\|_{F}
    \label{eq:st1}
\end{align}
with a minimum at the fixed point $V_{*} = span\{v_{1}, v_{2},\ldots,v_{r}\}$ where $v_{i=1,2,\ldots,r}$ are the right singular vectors of $X_{r}$ corresponding to the lowest singular values including zeros. This minimization problem describes the Stage 1 loss function of our network architecture. Upon convergence, the minimizer $\tilde{V}_{*}$ is such that $V_{*}V_{*}^{T} = V_{r}V_{r}^{T}$ where $V_{r}$ is the matrix with columns as right singular vectors of $X$ corresponding to the smallest $r$ singular values of $X$.

\begin{remark}
Note that the orthonormality constraint is necessary to avoid trivial solutions corresponding to $\tilde{V}=[v_1,\cdots,v_r]$ where $v_i \in null(X)$ $\forall i = 1,\cdots, r$ .
\end{remark}

The penalization $\alpha = \|X\|_{F}$ is chosen empirically for the time being pending further analysis. As before with Range-Net, Stage-1 of Tail-Net is a flexible module to extract rank-$r$ subspace of $X$ or identifying the rank of a system. For $r\leq (g-f)$, Stage-1 loss will always converges to a zero b-tail energy.

\textbf{Stage 2: Singular Value and Vector Extraction:} The rotation stage then extracts the singular values by rotating the orthonormal vectors ($V_{*}$) to align with the right singular vectors ($V_{r} = V_{*}\Theta_{r}$). From the fixed point of the Stage-1 minimization problem Eq. \eqref{eq:st1} we have $V_{*}V_{*}^{T} = V_{r}V_{r}^{T}$. According to the EYM theorem the tail energy of a rank-r matrix $XV_{*}C_{r}$, where $C_{r}$ is an arbitrary rank-r, real valued, square matrix, with respect to $XV_{*}$ is now bounded below by 0,
\begin{align*}
    \|XV_{*} - XV_{*}C_{r}\|_{F} \geq  0\\
\end{align*}
Borrowing from Range-Net, we know that $C_{r} = \Theta_{r}\Theta_{r}^{T}$, where $\Theta{r}$ is a rank-r, unitary matrix in an r-dimensional Euclidean space. Further, $(XV_{*}\Theta_{r})^{T}(XV_{*}\Theta_{r})$ is a diagonal matrix $\Sigma^{2}_{r} = \mathrm{diag}(\sigma^{2}_{1},\sigma^{2}_{2},\cdots,\sigma^{2}_{r})$, where $\sigma_{i}$s are the bottom-r singular values of $X$ if and only if $V_{*}\Theta_{r} = V_{r}$. Assuming  $Y =  XV_{*}\Theta_{r}$ for convenience of notation, the minimization problem reads:
\begin{align*}
\underset{\Theta_{r}}{\min} \quad  & \|Y\Theta_{r}^{T} - XV_{*}\|_{F}\\
 & s.t.  \text{ and } \quad Y^{T}Y - \mathrm{diag}(Y^{T}Y) = 0\\
\end{align*} 

\begin{remark}
Note that for Tail-Net, the stage-2 minimization problem remains unchanged when compared to the originally proposed Range-Net \cite{singh2021range}.
\end{remark}
As discussed previously, this choice of loss terms equipped with a Frobenius norm ensures a rank-$r$ approximation in accord with the Eckart-Young-Mirsky (EYM) theorem. We are therefore able to preemptively state that the expected value of the stage-1 loss term at the minimum corresponds to the rank $r$ b-tail energy. This can be verified by performing a full SVD using conventional solvers and computing a Frobenius norm on a reconstruction of the data using the bottom $(r)$ singular values and vectors. Further, the second loss term/ tail energy is expected to reach a machine precision zero at the minimum. Once, the network minimization problem converges, the singular values are extracted from \textbf{Stage 2} network weights $\Theta_{r}$ as $\Sigma_{r}^{2} = (XV_{*}\Theta_{r})^{T}(XV_{*}\Theta_{r})$. The right singular vectors can now be extracted using \textbf{Stage 2} layer weights given by $V_{r} = V_{*}\Theta_{r}$. Once $V_{r}$ and $\Sigma_{r}$ are known, left singular vectors $U_{r} = XV_{*}\Theta_{r}\Sigma_{r}^{-1}$. Please note that for $r>f$, $f-r$ singular values are zero and therefore $\Sigma_{r}^{-1}$ implies inverting the non-zero singular values that exceed a threshold of $\epsilon = 10^{-8}$.

\subsection{Network Interpretability} \label{sec:inter}

As described in before Fig. \ref{fig:arch}, our network weights and outputs are strictly defined with appropriate losses in the network minimization problem. In order to create a distinction, we refer to the problem informed (SVD) restrictions on the network weights as representation driven losses. The representation driven, orthonormality loss term, in Stage 1 enforces $\tilde{V}$ to be orthonormal or $(V_{*}^TV_{*}=I_r)$ for a desired rank-$r$ lowest singular triplets. We numerically verify the interpretability of the layer outputs and weights by considering two networks: (1) with, and (2) without the aforementioned orthonomality loss. For each of these two cases, three synthetic datasets are considered corresponding to $f = 10, 7, 5$ where the lowest $5$ singular vectors are desired. Note that in a practical scenario $f$ is an unknown and can be determined only by performing a full SVD of $X$.

\begin{figure}[h]
    \centering
    \includegraphics[width=0.8\linewidth]{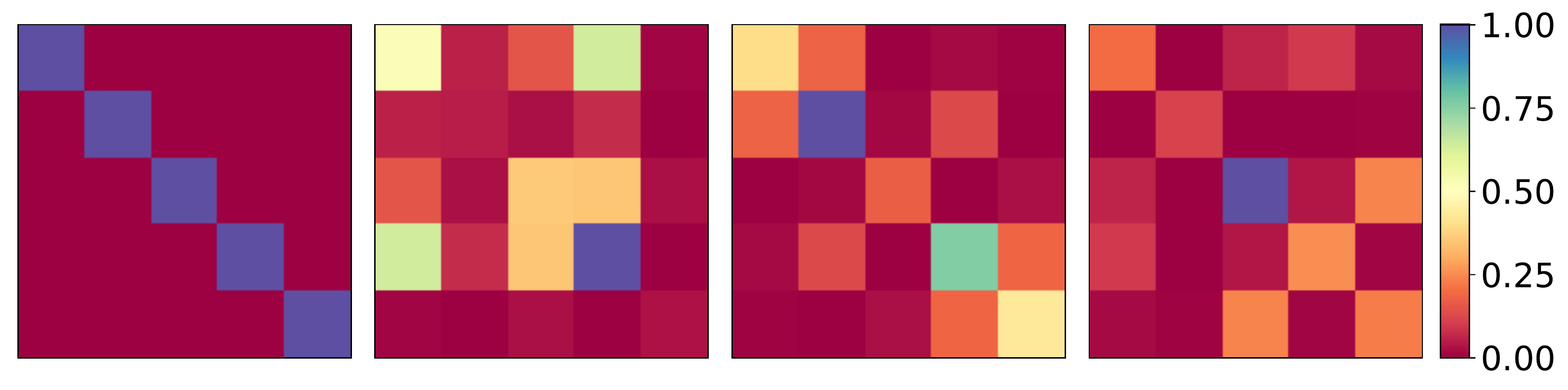}
    \caption{Synthetic Full Rank Scenario: Correlation Map of extracted vectors $V_{*}$ over four runs. Orthonormality is imposed for the first run resulting in a diagonal structure. Absence of this condition results in a scatter for the remaining three runs, as expected.}
    \label{fig:synth1}
\end{figure}

For the first case, we consider a full-rank synthetic data matrix $X_{10 \times 10}$ where ($f=10$). The objective is to extract the bottom $5$ ($r=5$) singular vectors. A total of four training runs are considered: one run for a network with the orthonormality condition imposed and three runs for another network without this additional constraints. \textbf{Fig. \ref{fig:synth1}} shows the correlation map between the recovered vectors $V_{*}$ for each of the four runs. Notice that only when the orthonormality criteria is not imposed, we get scatter away from the diagonal matrix, although all four runs converged to the same tail energy. The absence of this orthonormality imposing representation loss results in non-orthonormal vectors $V_{*}$.

\begin{figure}[h]
    \centering
    \includegraphics[width=0.8\linewidth]{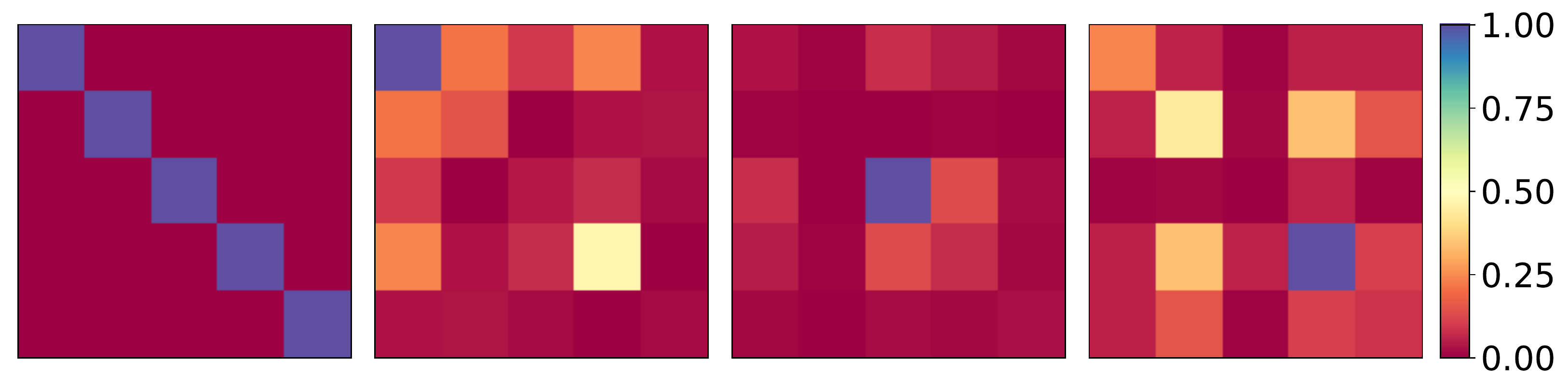}
    \caption{Synthetic Medium rank Scenario: Correlation Map of extracted vectors $\tilde{V}$ over four runs. First one with orthonormality loss imposed, rest three without it.}
    \label{fig:synth2}
\end{figure}

For the second case, we consider a low-rank synthetic data matrix $X_{10 \times 10}$ with $f=7$. Under this scenario we have $2$ non-zero and $3$ zero singular values. As before, we extract the bottom $5$ singular vectors ($r=5$) using four training runs: one with and three without imposing the orthonormality loss. Fig. \textbf{Fig. \ref{fig:synth2}} shows that the extracted vectors $V_{*}$ remain orthonormal only for the first run to minimize the b-tail energy ($\|X\tilde{V}\tilde{V}^T\|_{F}$), as described in \textbf{Section \ref{sec:arch}} above.

\begin{figure}[h]
    \centering
    \includegraphics[width=0.8\linewidth]{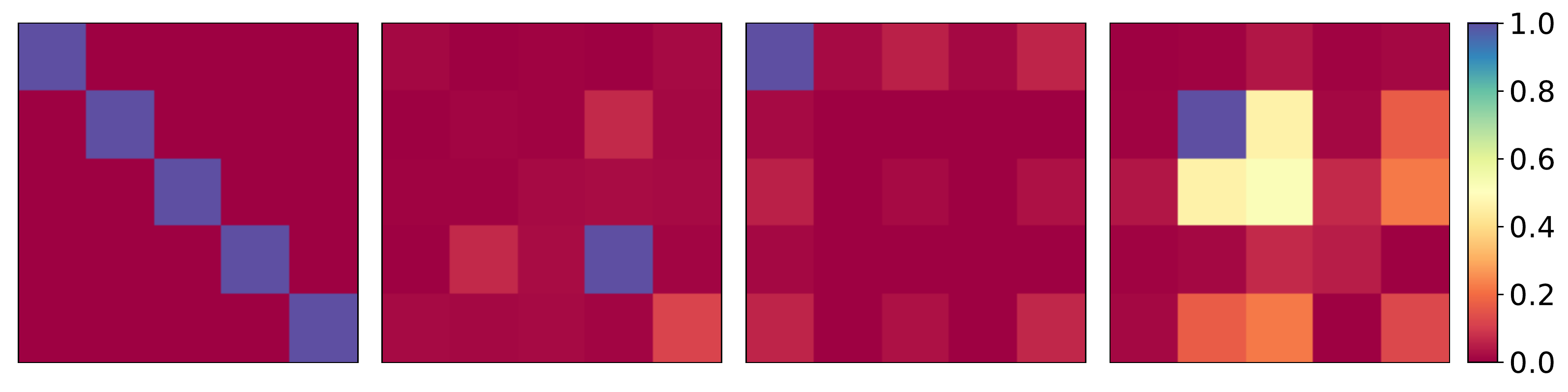}
    \caption{Synthetic Low Rank Scenario: Correlation Map of extracted vectors $\tilde{V}$ over four runs. First one with orthonormality loss imposed, rest three without it.}
    \label{fig:synth3}
\end{figure}

For the third case, we consider another low-rank synthetic data matrix $X_{10 \times 10}$ with $f=5$. As before, we extract the bottom $5$ singular vectors ($r=5$) corresponding to the $5$ zero singular values using four training runs: one with and three without imposing the orthonormality loss. Note that in this scenario the b-tail energy converges to GPU precision zero since all the $5$ singular values are zero. Fig. \textbf{Fig. \ref{fig:synth3}} shows that the extracted vectors $V_{*}$ remain orthonormal only for the first run. 

\section{Results}
We begin this section by first defining error metrics for comparison and bench-marking purposes. In the subsequent subsections, we present our training setup, results and analysis for various synthetic and real datasets, which vary in scale from small to big data. Please note that our numerical results do not show large variations in these error metrics over multiple runs due to a precise low-weight architecture. For all of the following numerical experiments, the \textbf{Stage 1} of our neural SVD solver requires at most 5 passes (empirical observation) over the data matrix to converge. Further, the error metrics rely upon conventional SVD as the baseline for a fair comparison.

\subsection{Metrics} \label{sec:metric}
We rely upon the following error metrics on the extracted factors for performance comparison and benchmarking. In the following $X$ and $\hat{X}$ are used to denote the true and the reconstructed data matrices.
\begin{itemize}
    \item \textbf{Reconstruction Error:} Frobenius norm of the element wise error of the true data and its rank-$r$ approximation.
    \begin{align*}
        frob_{err}(r) = \|X - \hat{X}\|_F^2 - \|X - X_r\|_F^2
    \end{align*}  
    \item \textbf{Spectral Error}: 2-norm of the singular value of the true data and its rank-$r$ approximation.
    \begin{align*}
        spectral_{err}(r) = \|X-\hat{X}\|_2 - \|X-X_r\|_2 
    \end{align*}
\end{itemize}
Here, $\sigma_{i}$s are the true singular values and $X_{r}$ is the desired lowest rank-$r$ approximation of $X$ using conventional SVD as the baseline for benchmarking. Under perfect recovery, all the error metrics are expected to approximately achieve zero at machine precision. All of our numerical experiments were performed on a GPU using single (32-bit) precision floating point operations. Therefore, the b-tail energies are expected to be correct to upto 8 significant digits approximately.

\subsection{Setup and Training}

All experiments were done on a setup with Nvidia 2060 RTX Super 8GB GPU, Intel Core i7-9700F 3.0GHz 8-core CPU and 16GB DDR4 memory. We use Keras \cite{chollet2015} library running on a Tensorflow 2.0 backend with Python 3.7 to train the networks presented in this paper. For optimization, we use AdaMax \cite{kingma2014adam} with parameters (\textit{lr}= 0.001) and $2000$ steps per epoch. The batch-sizes vary with dataset sizes and are therefore not reported explicitly.

\subsection{Real Network Graph: Wiki-Vote} \label{sec:netgraph}

\begin{figure}[h]
    \centering
    \begin{subfigure}{.45\linewidth}
      \centering
      \includegraphics[width=0.6\linewidth]{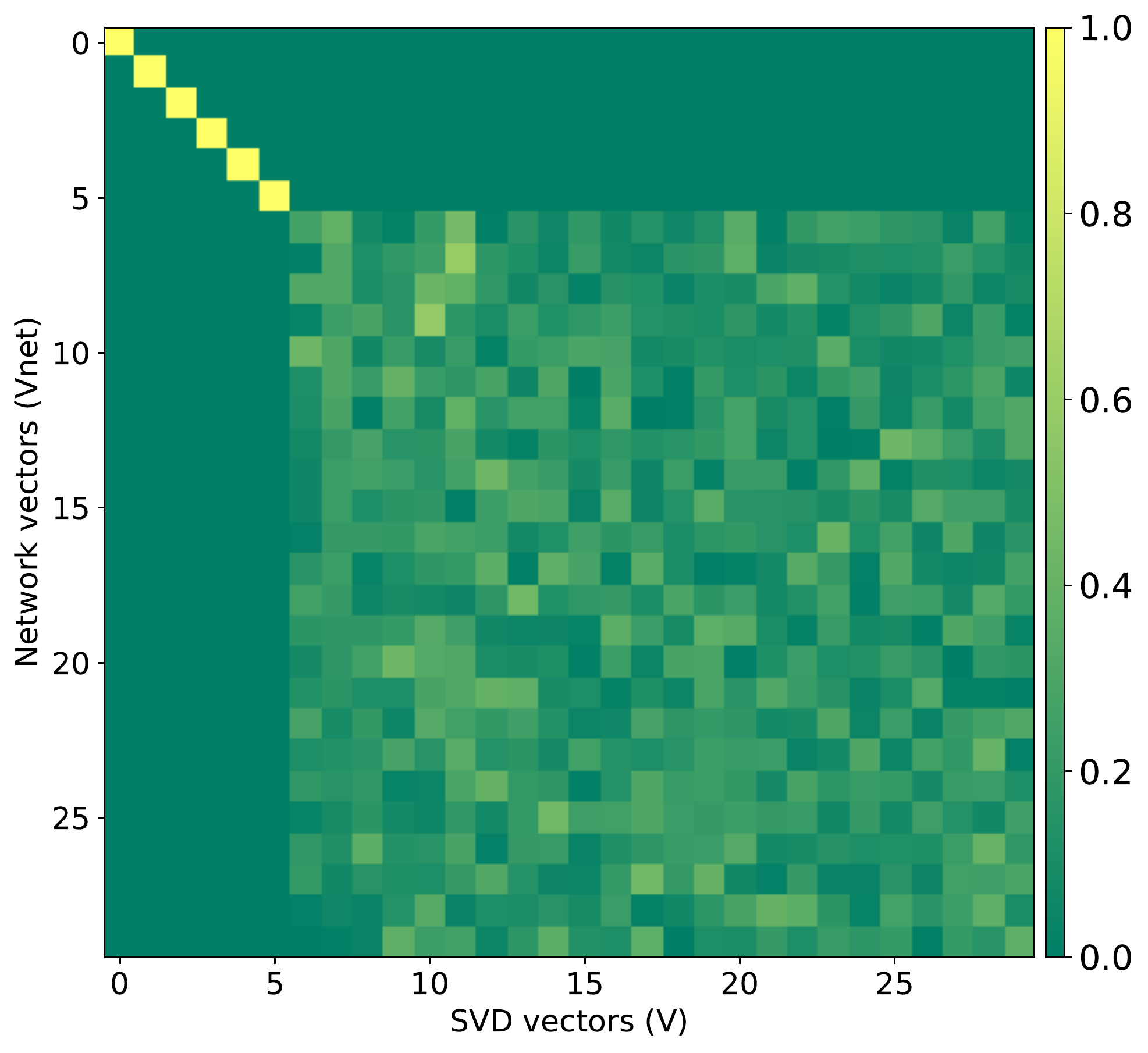}  
      \caption{Cross-correlation between extracted and true singular vectors}
    \end{subfigure}
    \begin{subfigure}{.45\linewidth}
      \centering
      \includegraphics[width=0.8\linewidth]{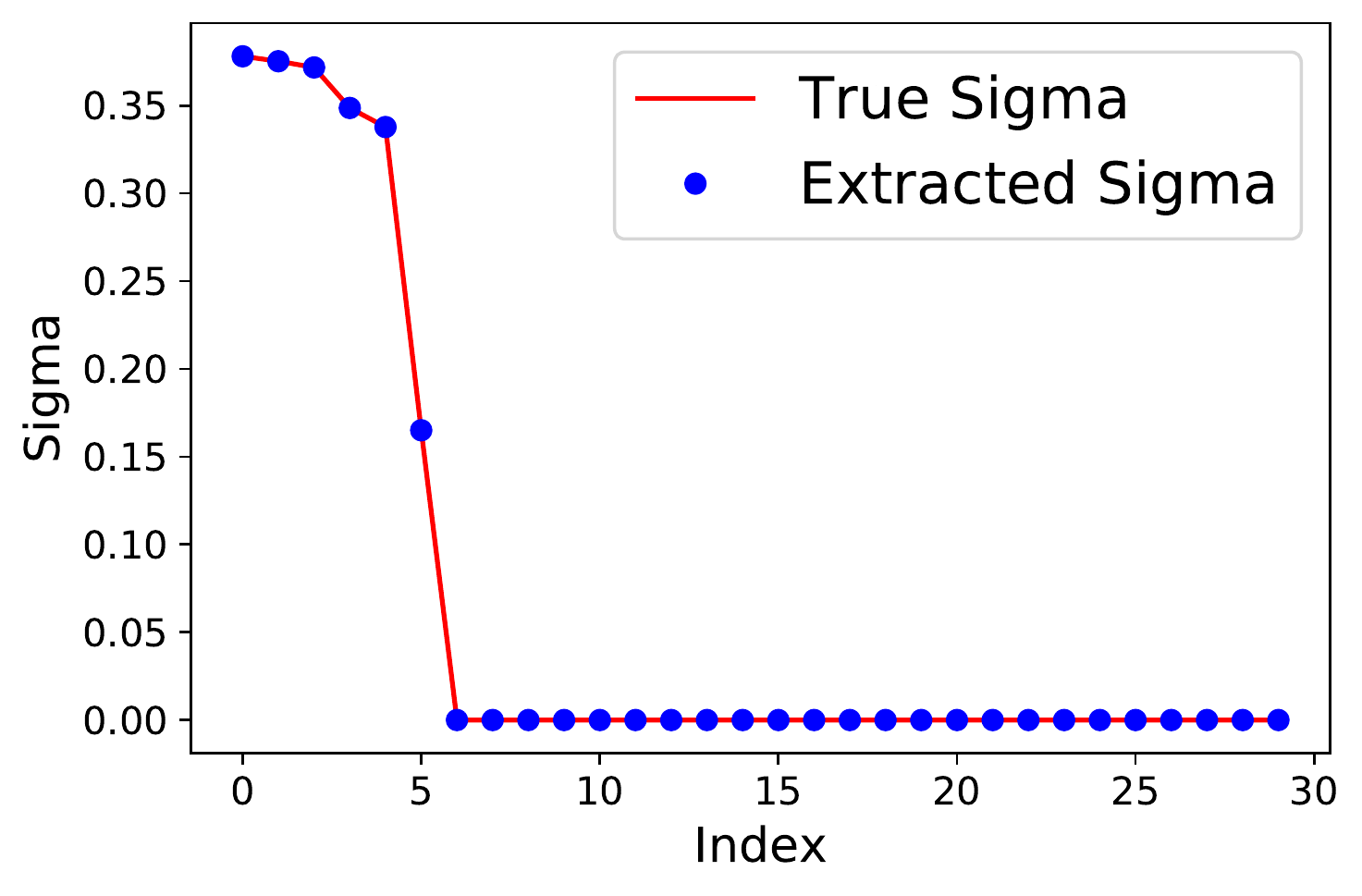}  
      \caption{Singular value spectrum}
    \end{subfigure}
    \caption{Extracting the lowest $r=30$ singular triplets (increasing order of singular values) from the graph Laplacian of Wikivote dataset. The lowest $24$ singular values are zero with $6$ non-zero singular values. The extracted triplets are compared against singular triplets from conventional SVD for comparison.}
    \label{fig:wiki}
\end{figure}

In this section, we consider the Wiki-Vote graph available in SNAP \cite{snapnets} Dataset. The graph contains $7115$ connected nodes and $103689$ edges over a directed acyclic graph. To achieve symmetry, we considered an un-directed version of its adjacency list. The numerical experiment was performed for extracting the lowest $30$ singular triplets on the graph Laplacian of the Wikivote adjacency matrix. Both conventional SVD and Tail-Net reveal $24$ zero singular values indicating $24$ isolated clusters with no edges connecting the individual clusters. \textbf{Fig. \ref{fig:wiki} (a)} shows the cross-correlation between singular vectors extracted using conventional SVD and Tail-Net. Note that the cross-correlation shows a scatter for the vectors corresponding to the zero singular values which is expected as these correspond to the null space of the data matrix.  Similarly, \textbf{Fig. \ref{fig:wiki} (b)} shows the singular value spectra from the two approaches for comparison and benchmarking. For the extracted singular triplets, the reconstruction and spectral error are $0.0$ and $0.0$ respectively. 

As claimed before, Tail-Net is rank revealing and is capable of handling zero singular values. For a practical dataset $X \in \mathbb{R}^{m \times n}$, the rank of the system is more likely to be close to $min(m,n)$ and therefore for rank estimation using Tail-Net is relatively easy. \textit{It is easier to chase and verify a zero b-tail energy than an unknown positive b-tail energy.} The stage-1 minimization problem of Tail-Net converges to the b-tail energy at GPU precision and can be verified against a conventional SVD scheme if feasible. The stage-2 minimization problem of Tail-Net converges to zero at GPU precision as in the case of Range-Net. Although Tail-Net can handle much larger  graphs, we restricted ourselves to adjacency matrix sizes where conventional SVD is viable for benchmarking purposes.

\section{Conclusion}

We present Tail-Net as a high-precision, fully interpretable SVD solver for extracting lowest singular triplets of a big data matrix. Our numerical experiments on real and synthetic datasets confirm that Tail-Net achieves the lower bound on the b-tail energy corresponding to the tail-energy proposed in the EYM theorem. We verify that our network minimization problems converges to this b-tail energy equipped with Frobenius norm at machine precision. We show numerical experiments on graph Laplacians of synthetic and real data adjacency matrices to demonstrate the applicability of Tail-Net to large scale practical datasets. A comparison is also provided against conventional SVD solvers for computational benchmarking and verification.

%
%

\begin{thebibliography}{20}
	
	
	\ifx \showCODEN    \undefined \def \showCODEN     #1{\unskip}     \fi
	\ifx \showDOI      \undefined \def \showDOI       #1{#1}\fi
	\ifx \showISBNx    \undefined \def \showISBNx     #1{\unskip}     \fi
	\ifx \showISBNxiii \undefined \def \showISBNxiii  #1{\unskip}     \fi
	\ifx \showISSN     \undefined \def \showISSN      #1{\unskip}     \fi
	\ifx \showLCCN     \undefined \def \showLCCN      #1{\unskip}     \fi
	\ifx \shownote     \undefined \def \shownote      #1{#1}          \fi
	\ifx \showarticletitle \undefined \def \showarticletitle #1{#1}   \fi
	\ifx \showURL      \undefined \def \showURL       {\relax}        \fi
	\providecommand\bibfield[2]{#2}
	\providecommand\bibinfo[2]{#2}
	\providecommand\natexlab[1]{#1}
	\providecommand\showeprint[2][]{arXiv:#2}
	
	\bibitem[\protect\citeauthoryear{Baglama and Reichel}{Baglama and
		Reichel}{2005}]%
	{baglama2005augmented}
	\bibfield{author}{\bibinfo{person}{James Baglama} {and} \bibinfo{person}{Lothar
			Reichel}.} \bibinfo{year}{2005}\natexlab{}.
	\newblock \showarticletitle{Augmented implicitly restarted Lanczos
		bidiagonalization methods}.
	\newblock \bibinfo{journal}{\emph{SIAM Journal on Scientific Computing}}
	\bibinfo{volume}{27}, \bibinfo{number}{1} (\bibinfo{year}{2005}),
	\bibinfo{pages}{19--42}.
	\newblock
	
	
	\bibitem[\protect\citeauthoryear{Bj{\"o}rck, Heggernes, and
		Matstoms}{Bj{\"o}rck et~al\mbox{.}}{2000}]%
	{bjorck2000methods}
	\bibfield{author}{\bibinfo{person}{{\AA}ke Bj{\"o}rck}, \bibinfo{person}{Pinar
			Heggernes}, {and} \bibinfo{person}{Pontus Matstoms}.}
	\bibinfo{year}{2000}\natexlab{}.
	\newblock \showarticletitle{Methods for large scale total least squares
		problems}.
	\newblock \bibinfo{journal}{\emph{SIAM J. Matrix Anal. Appl.}}
	\bibinfo{volume}{22}, \bibinfo{number}{2} (\bibinfo{year}{2000}),
	\bibinfo{pages}{413--429}.
	\newblock
	
	
	\bibitem[\protect\citeauthoryear{Bradley and Mangasarian}{Bradley and
		Mangasarian}{2000}]%
	{bradley2000k}
	\bibfield{author}{\bibinfo{person}{Paul~S Bradley} {and}
		\bibinfo{person}{Olvi~L Mangasarian}.} \bibinfo{year}{2000}\natexlab{}.
	\newblock \showarticletitle{K-plane clustering}.
	\newblock \bibinfo{journal}{\emph{Journal of Global Optimization}}
	\bibinfo{volume}{16}, \bibinfo{number}{1} (\bibinfo{year}{2000}),
	\bibinfo{pages}{23--32}.
	\newblock
	
	
	\bibitem[\protect\citeauthoryear{Chollet}{Chollet}{2015}]%
	{chollet2015}
	\bibfield{author}{\bibinfo{person}{François Chollet}.}
	\bibinfo{year}{2015}\natexlab{}.
	\newblock \bibinfo{title}{keras}.
	\newblock \bibinfo{howpublished}{\url{https://github.com/fchollet/keras}}.
	\newblock
	
	
	\bibitem[\protect\citeauthoryear{Dax}{Dax}{2019}]%
	{dax2019computing}
	\bibfield{author}{\bibinfo{person}{Achiya Dax}.}
	\bibinfo{year}{2019}\natexlab{}.
	\newblock \showarticletitle{Computing the smallest singular triplets of a large
		matrix}.
	\newblock \bibinfo{journal}{\emph{Results in Applied Mathematics}}
	\bibinfo{volume}{3} (\bibinfo{year}{2019}), \bibinfo{pages}{100006}.
	\newblock
	
	
	\bibitem[\protect\citeauthoryear{Eckart and Young}{Eckart and Young}{1936}]%
	{eckart1936approximation}
	\bibfield{author}{\bibinfo{person}{Carl Eckart} {and} \bibinfo{person}{Gale
			Young}.} \bibinfo{year}{1936}\natexlab{}.
	\newblock \showarticletitle{The approximation of one matrix by another of lower
		rank}.
	\newblock \bibinfo{journal}{\emph{Psychometrika}} \bibinfo{volume}{1},
	\bibinfo{number}{3} (\bibinfo{year}{1936}), \bibinfo{pages}{211--218}.
	\newblock
	
	
	\bibitem[\protect\citeauthoryear{Elsner and He}{Elsner and He}{1991}]%
	{elsner1991algorithm}
	\bibfield{author}{\bibinfo{person}{Ludwig Elsner} {and}
		\bibinfo{person}{Chunyang He}.} \bibinfo{year}{1991}\natexlab{}.
	\newblock \showarticletitle{An algorithm for computing the distance to
		uncontrollability}.
	\newblock \bibinfo{journal}{\emph{Systems \& control letters}}
	\bibinfo{volume}{17}, \bibinfo{number}{6} (\bibinfo{year}{1991}),
	\bibinfo{pages}{453--464}.
	\newblock
	
	
	\bibitem[\protect\citeauthoryear{Hardoon, Szedmak, and Shawe-Taylor}{Hardoon
		et~al\mbox{.}}{2004}]%
	{hardoon2004canonical}
	\bibfield{author}{\bibinfo{person}{David~R Hardoon}, \bibinfo{person}{Sandor
			Szedmak}, {and} \bibinfo{person}{John Shawe-Taylor}.}
	\bibinfo{year}{2004}\natexlab{}.
	\newblock \showarticletitle{Canonical correlation analysis: An overview with
		application to learning methods}.
	\newblock \bibinfo{journal}{\emph{Neural computation}} \bibinfo{volume}{16},
	\bibinfo{number}{12} (\bibinfo{year}{2004}), \bibinfo{pages}{2639--2664}.
	\newblock
	
	
	\bibitem[\protect\citeauthoryear{Kingma and Ba}{Kingma and Ba}{2014}]%
	{kingma2014adam}
	\bibfield{author}{\bibinfo{person}{Diederik~P Kingma} {and}
		\bibinfo{person}{Jimmy Ba}.} \bibinfo{year}{2014}\natexlab{}.
	\newblock \showarticletitle{Adam: A method for stochastic optimization}.
	\newblock \bibinfo{journal}{\emph{arXiv preprint arXiv:1412.6980}}
	(\bibinfo{year}{2014}).
	\newblock
	
	
	\bibitem[\protect\citeauthoryear{Larsen}{Larsen}{1998}]%
	{larsen1998lanczos}
	\bibfield{author}{\bibinfo{person}{Rasmus~Munk Larsen}.}
	\bibinfo{year}{1998}\natexlab{}.
	\newblock \showarticletitle{Lanczos bidiagonalization with partial
		reorthogonalization}.
	\newblock \bibinfo{journal}{\emph{DAIMI Report Series}} \bibinfo{number}{537}
	(\bibinfo{year}{1998}).
	\newblock
	
	
	\bibitem[\protect\citeauthoryear{Leskovec and Krevl}{Leskovec and
		Krevl}{2014}]%
	{snapnets}
	\bibfield{author}{\bibinfo{person}{Jure Leskovec} {and} \bibinfo{person}{Andrej
			Krevl}.} \bibinfo{year}{2014}\natexlab{}.
	\newblock \bibinfo{title}{{SNAP Datasets}: {Stanford} Large Network Dataset
		Collection}.
	\newblock \bibinfo{howpublished}{\url{http://snap.stanford.edu/data}}.
	\newblock
	
	
	\bibitem[\protect\citeauthoryear{Luo, Unbehauen, and Cichocki}{Luo
		et~al\mbox{.}}{1997}]%
	{luo1997minor}
	\bibfield{author}{\bibinfo{person}{Fa-Long Luo}, \bibinfo{person}{Rolf
			Unbehauen}, {and} \bibinfo{person}{Andrzej Cichocki}.}
	\bibinfo{year}{1997}\natexlab{}.
	\newblock \showarticletitle{A minor component analysis algorithm}.
	\newblock \bibinfo{journal}{\emph{Neural Networks}} \bibinfo{volume}{10},
	\bibinfo{number}{2} (\bibinfo{year}{1997}), \bibinfo{pages}{291--297}.
	\newblock
	
	
	\bibitem[\protect\citeauthoryear{Mirsky}{Mirsky}{1960}]%
	{mirsky1960symmetric}
	\bibfield{author}{\bibinfo{person}{Leon Mirsky}.}
	\bibinfo{year}{1960}\natexlab{}.
	\newblock \showarticletitle{Symmetric gauge functions and unitarily invariant
		norms}.
	\newblock \bibinfo{journal}{\emph{The quarterly journal of mathematics}}
	\bibinfo{volume}{11}, \bibinfo{number}{1} (\bibinfo{year}{1960}),
	\bibinfo{pages}{50--59}.
	\newblock
	
	
	\bibitem[\protect\citeauthoryear{Reiss, Thomas, and Reiss}{Reiss
		et~al\mbox{.}}{1997}]%
	{reiss1997statistical}
	\bibfield{author}{\bibinfo{person}{Rolf-Dieter Reiss}, \bibinfo{person}{Michael
			Thomas}, {and} \bibinfo{person}{RD Reiss}.} \bibinfo{year}{1997}\natexlab{}.
	\newblock \bibinfo{booktitle}{\emph{Statistical analysis of extreme values}}.
	Vol.~\bibinfo{volume}{2}.
	\newblock \bibinfo{publisher}{Springer}.
	\newblock
	
	
	\bibitem[\protect\citeauthoryear{Singh, Gupta, Lease, and Dawson}{Singh
		et~al\mbox{.}}{2021}]%
	{singh2021range}
	\bibfield{author}{\bibinfo{person}{Gurpreet Singh}, \bibinfo{person}{Soumyajit
			Gupta}, \bibinfo{person}{Matthew Lease}, {and} \bibinfo{person}{Clint
			Dawson}.} \bibinfo{year}{2021}\natexlab{}.
	\newblock \showarticletitle{Range-Net: A High Precision Streaming SVD for Big
		Data Applications}.
	\newblock \bibinfo{journal}{\emph{arXiv preprint arXiv:2010.14226}}
	(\bibinfo{year}{2021}).
	\newblock
	
	
	\bibitem[\protect\citeauthoryear{Trefethen}{Trefethen}{1999}]%
	{trefethen1999computation}
	\bibfield{author}{\bibinfo{person}{Lloyd~N Trefethen}.}
	\bibinfo{year}{1999}\natexlab{}.
	\newblock \showarticletitle{Computation of pseudospectra}.
	\newblock \bibinfo{journal}{\emph{Acta Numerica}}  \bibinfo{volume}{8}
	(\bibinfo{year}{1999}), \bibinfo{pages}{247--295}.
	\newblock
	
	
	\bibitem[\protect\citeauthoryear{Van Der~Veen, Deprettere, and
		Swindlehurst}{Van Der~Veen et~al\mbox{.}}{1993}]%
	{van1993subspace}
	\bibfield{author}{\bibinfo{person}{A-J Van Der~Veen}, \bibinfo{person}{ED~F
			Deprettere}, {and} \bibinfo{person}{A~Lee Swindlehurst}.}
	\bibinfo{year}{1993}\natexlab{}.
	\newblock \showarticletitle{Subspace-based signal analysis using singular value
		decomposition}.
	\newblock \bibinfo{journal}{\emph{Proc. IEEE}} \bibinfo{volume}{81},
	\bibinfo{number}{9} (\bibinfo{year}{1993}), \bibinfo{pages}{1277--1308}.
	\newblock
	
	
	\bibitem[\protect\citeauthoryear{Von~Luxburg}{Von~Luxburg}{2007}]%
	{von2007tutorial}
	\bibfield{author}{\bibinfo{person}{Ulrike Von~Luxburg}.}
	\bibinfo{year}{2007}\natexlab{}.
	\newblock \showarticletitle{A tutorial on spectral clustering}.
	\newblock \bibinfo{journal}{\emph{Statistics and computing}}
	\bibinfo{volume}{17}, \bibinfo{number}{4} (\bibinfo{year}{2007}),
	\bibinfo{pages}{395--416}.
	\newblock
	
	
	\bibitem[\protect\citeauthoryear{Wiskott and Sejnowski}{Wiskott and
		Sejnowski}{2002}]%
	{wiskott2002slow}
	\bibfield{author}{\bibinfo{person}{Laurenz Wiskott} {and}
		\bibinfo{person}{Terrence~J Sejnowski}.} \bibinfo{year}{2002}\natexlab{}.
	\newblock \showarticletitle{Slow feature analysis: Unsupervised learning of
		invariances}.
	\newblock \bibinfo{journal}{\emph{Neural computation}} \bibinfo{volume}{14},
	\bibinfo{number}{4} (\bibinfo{year}{2002}), \bibinfo{pages}{715--770}.
	\newblock
	
	
	\bibitem[\protect\citeauthoryear{Wu and Stathopoulos}{Wu and
		Stathopoulos}{2014}]%
	{wu2014primme}
	\bibfield{author}{\bibinfo{person}{Lingfei Wu} {and} \bibinfo{person}{Andreas
			Stathopoulos}.} \bibinfo{year}{2014}\natexlab{}.
	\newblock \showarticletitle{Primme svds: A preconditioned svd solver for
		computing accurately singular triplets of large matrices based on the primme
		eigensolver}.
	\newblock \bibinfo{journal}{\emph{arXiv preprint arXiv:1408.5535}}
	(\bibinfo{year}{2014}).
	\newblock
	
	
\end{thebibliography}

%
\appendix

\end{document}